%% file: acl_latex.tex
\pdfoutput=1

\documentclass[11pt]{article}

\usepackage[]{acl}
\usepackage{epsfig}
\usepackage{times}
\usepackage{latexsym}
\usepackage{algorithm}
\usepackage{algorithmic}
\usepackage[misc]{ifsym} 

\usepackage[T1]{fontenc}

\usepackage[utf8]{inputenc}

\usepackage{microtype}
\usepackage{arydshln} 

\usepackage{inconsolata}

\title{\textbf{XAL}: \textbf{EX}plainable \textbf{A}ctive \textbf{L}earning Makes Classifiers Better Low-resource Learners}

\author{
\centerline{Yun Luo$^{1,3}$ ~~ Zhen Yang$^{2}$ ~~ Fandong Meng$^{2}$ ~~ Yingjie Li $^{1}$ ~~ Fang Guo$^{1,3}$}   \\
\centerline{ \textbf{Qinglin Qi}$^4$ ~~ \textbf{Jie Zhou}$^2$ ~~ \textbf{Yue Zhang}$^{1,5, \textrm{\small\Letter}}$} \\
 \centerline{\normalfont{$^1$School of Engineering, Westlake University}  \quad \quad\normalfont{$^2$WeChat AI, Tencent Inc.}}\\
 \centerline{\normalfont{$^3$ Zhejiang University} \normalfont{$^4$ School of Cyber Science and Engineering, Sichuan University}}\\
\centerline{\normalfont{$^5$ Institute of Advanced Technology, Westlake Institute for Advanced Study}}\\
\centerline{\texttt{\{luoyun,zhangyue\}@westlake.edu.cn}}
}

\begin{document}
\maketitle
\begin{abstract}
Active learning (AL), which aims to construct an effective training set by iteratively curating the most formative unlabeled data for annotation, has been widely used in low-resource tasks. Most active learning techniques in classification rely on the model's uncertainty or disagreement to choose unlabeled data, suffering from the problem of 
 over-confidence in superficial patterns and a lack of exploration.
Inspired by the cognitive processes in which humans deduce and predict through causal information, we take an initial attempt towards integrating rationales into AL and propose a novel Explainable Active Learning framework (XAL) for low-resource text classification, which aims to encourage classifiers to justify their inferences and delve into unlabeled data for which they cannot provide reasonable explanations.  Specifically, besides using a pre-trained bi-directional encoder for classification, we employ a pre-trained uni-directional decoder to generate and score the explanation. 
We further facilitate the alignment of the model with human reasoning preference through a proposed ranking loss.
During the selection of unlabeled data, the predicted uncertainty of the encoder and the explanation score of the decoder complement each other as the final metric to acquire informative data. Extensive experiments on six datasets show that XAL achieves consistent improvement over 9 strong baselines. Analysis indicates that the proposed method can generate corresponding explanations for its predictions.



\end{abstract}

\input{intro}

\input{related}

\input{method}

\input{exp}

\input{results}

\input{conclusion}

\bibliography{custom}

\appendix
\input{appendix}

\end{document}

%% file: intro.tex
\section{Introduction}
Active learning (AL) is a machine-learning paradigm that efficiently acquires data for annotation from a (typically large) unlabeled data pool and iteratively trains models \citep{lewis1994heterogeneous,margatina2021active}. AL frameworks have attracted considerable attention from researchers due to their high realistic values reduce the data annotation costs by concentrating the human labeling effort on the most informative data points, which can be applied in low-resources tasks \citep{lewis1994heterogeneous,settles2009active,zhangsurvey}.

Most previous AL methods rely on model predictive uncertainty or disagreement for unlabeled data, and the most uncertain data are believed to be the most informative and worthful ones to be annotated \citep{lewis1995sequential,houlsby2011bayesian,margatina2021active,zhang_2022}. However, previous studies have indicated that existing models struggle to accurately quantify predictive uncertainty \citep{guo2017calibration,lakshminarayanan2017simple}, leading to overconfidence and insufficient exploration, i.e., models tend to choose data instances that are uncertain yet repetitively uninformative \citep{margatina2021active}. This issue arises because training can lead cross-entropy-based classifiers to learn superficial or spurious patterns \citep{Yiduo,guo,srivastava2020robustness}, rather than the causal information between inputs and labels.

\begin{figure*}
    \centering
    \includegraphics[width=0.9\hsize]{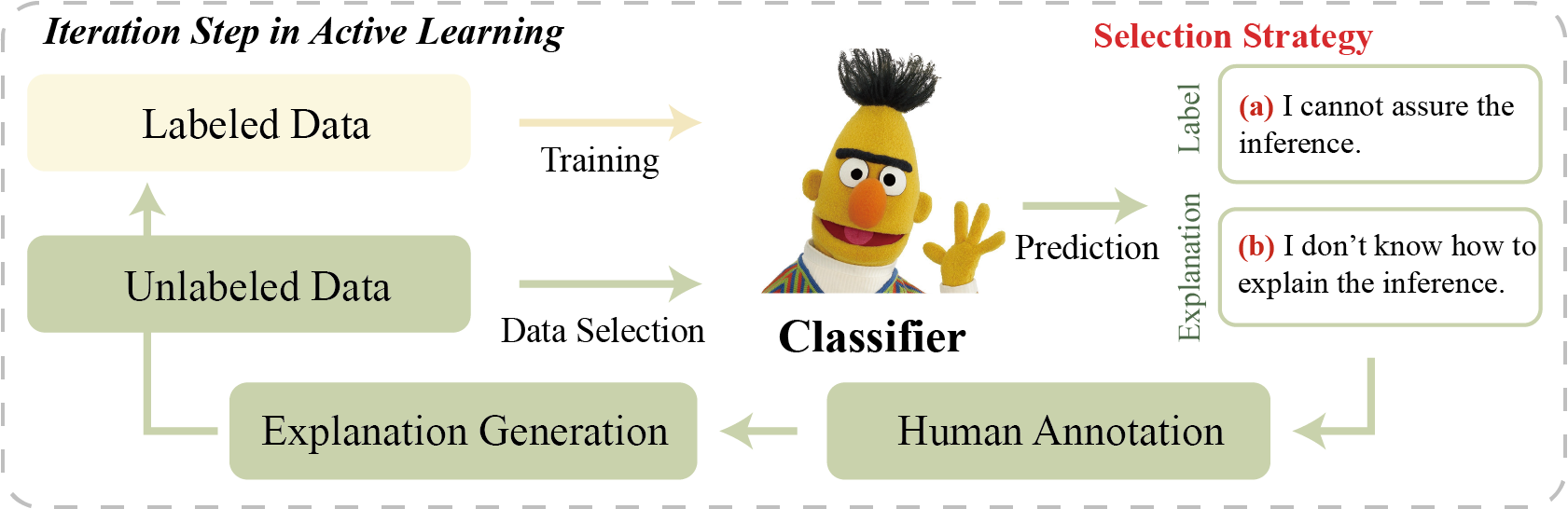}
    \vspace{-1mm}
    \caption{Data selection strategy in AL. Previous work selects the unlabeled data mostly relying on the model's uncertainly \textcolor{red}{(a)}, but we propose to further leverage the model's explanation of its prediction \textcolor{red}{(b)}.}
    \label{intro_f}
    \vspace{-3mm}
\end{figure*}

In the context of cognitive science and psychological science, 
humans make decisions or inferences by exploring causal information \citep{frye1996inference,joyce1999foundations,rottman2014reasoning}. 
For example, when learning to differentiate animals, humans do not merely rely on statistical features such as colors or feathers. They also consider the creatures' habits, such as dietary patterns, and kinship, such as the species of the parents, to engage in exploring rationales, thereby determining the species of the organism. 
Intuitively, explanations of the rational can help the model confirm whether it understands how to make classifications, and explaining the reasons behind the classification also enhances the justification of the inference confidence. It motivates us to encourage classifiers to learn the rationales behind inferences and explore unlabeled data for which the model cannot provide reasonable explanations. In doing so, the model can learn rationales between labels and texts and reduce reliance on superficial patterns, which leads to improved generalization and more effective exploration within AL frameworks. The intuition is illustrated in Figure \ref{intro_f}.

Given the above observations, we introduce an Explainable Active Learning Framework (XAL) for text classification tasks. 
This framework consists of two main components: the training process and the data selection process. Primarily, we adopt a pre-trained bi-directional encoder for classification and a pre-trained uni-directional decoder to generate and score explanations that serve as expressions of rationales in human language. In the training phase, we use the classification labels and explanations to optimize the model parameters. Besides, to further enhance the decoder's ability to score explanations, we design a ranking loss that optimizes the model to differentiate between reasonable and unreasonable explanations. To implement this ranking loss, we generate a variety of explanations (both reasonable and unreasonable) for labeled data by querying ChatGPT with different prompts, thereby eliminating the need for additional human annotation effort. Subsequently, during the data selection phase, we amalgamate the predictive uncertainty of the encoder and the explanation score of the decoder to rank unlabeled data. The most informative data are then annotated and incorporated into further training.

We conduct experiments on various text classification tasks involving different level of difficulty in understanding rationals. Experimental results manifest that XAL can achieve substantial improvement in all tasks. Ablation studies demonstrate the effectiveness of each component, and human evaluation shows that the model trained in XAL works well in explaining its prediction. XAL also demonstrates superior performance with only 500 instances when compared to in-context learning by ChatGPT, underscoring the effectiveness of our model at a minimal cost.
To our knowledge, we are the first to incorporate the model's explanation (explanation score) to improve the effectiveness of data selection in AL process. 
The codes and data have been released in the link to facilitate further research \footnote{\href{https://github.com/LuoXiaoHeics/XAL}{https://github.com/LuoXiaoHeics/XAL}}. 



%% file: related.tex
\section{Related Work}

\begin{figure*}
    \centering
    \includegraphics[width=0.86\hsize]{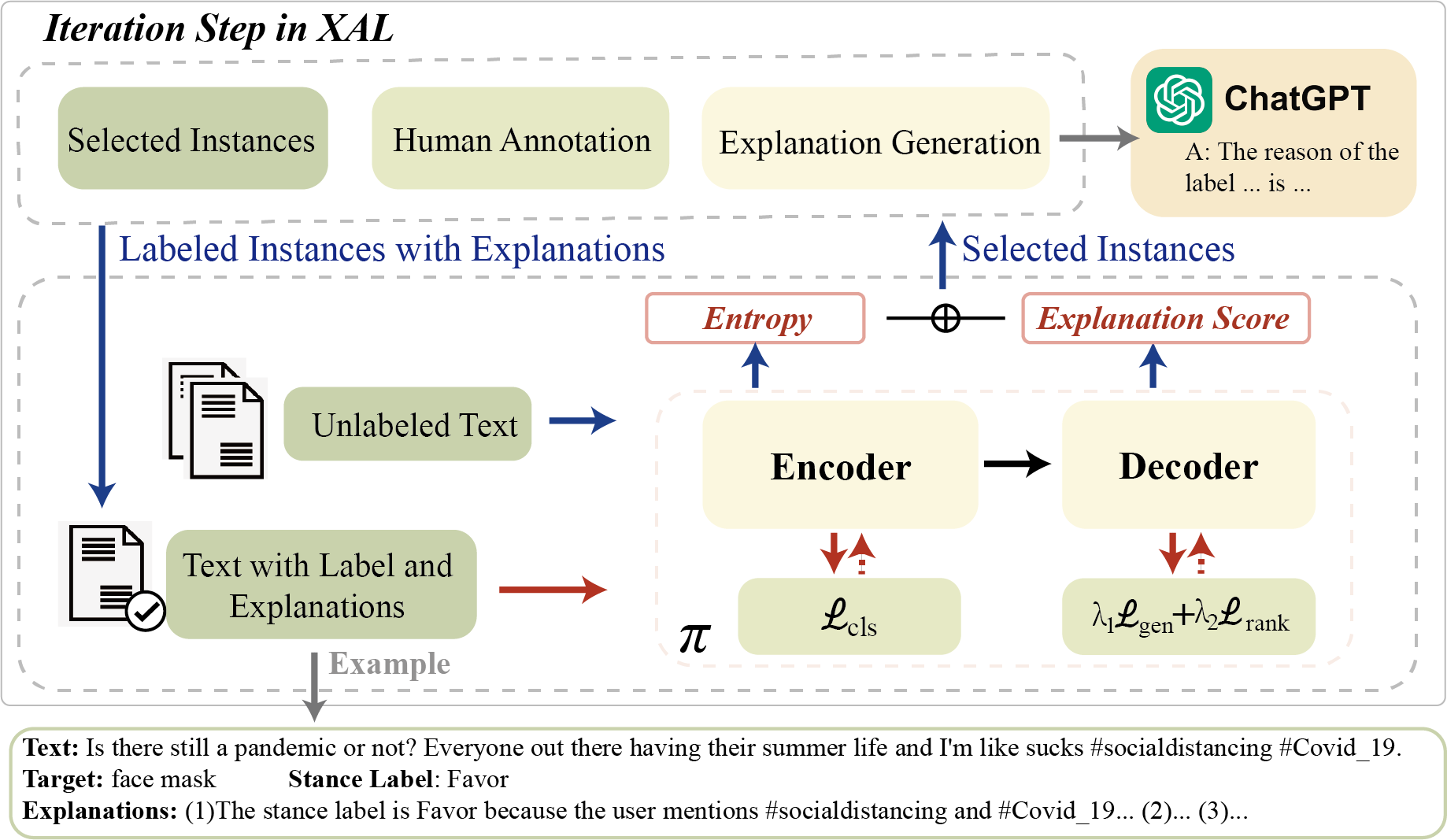}
    \vspace{-1mm}
    \caption{Our proposed XAL framework, which can be divided into two main parts -- the \textcolor{red}{training process} (red arrows) and the \textcolor{blue}{data selection process} (blue arrows). The training process aims to train the encoder-decoder model to learn classification and explanation generation. The data selection process aims to select unlabeled data using predictive entropy and explanation scores.  }
    \label{model}
    \vspace{-3mm}
\end{figure*}

\textbf{Active Learning}  is widely studied in the natural language processing area, ranging from text classification \citep{roy2001toward,zhang2017active,maekawa2022low}, and sequence labeling \citep{settlesanalysis} to text generation \citep{zhaoactive}. Previous methods can be roughly divided into informativeness-based selection strategies, representativeness-based selection strategies, and hybrid selection strategies \citep{zhang_2022}. The most mainstream methods, i.e., informativeness-based methods, are mostly characterized using model uncertainty, disagreement, or performance prediction, which suffers from over-confidence and a lack of exploration \citep{guo2017calibration,margatina2021active}. On the other hand, the representativeness-based methods rely on model inputs such as the representations of texts, which tends to select simple data samples and results in unsatisfactory performance \citep{roy2001toward,margatina2021active}.

\textbf{Large Language Model}. Recently, LLMs of generative schema have shown excellent performance in various NLP tasks \cite{bubeck2023sparks，yang2023teal}. However, some studies show that in-context learning based on LLMs \citep{radford2019language,brown2020language} suffers from practical issues such as high computation costs for inference \citep{liu2022few},  inclination to their internal knowledge \cite{yan2024refutebench}, catastrophic forgetting during instruction tuning \cite{luo2023empirical}, over-sensitive to example choices and instruction wording \citep{gao2020making,schick-schutze-2021-just}. Considering the problems, LLMs are also applied to improve the generalization of smaller models for specific tasks. Some studies distill the knowledge of large language models to a smaller one \cite{hsieh2023distilling} and some also use LLMs to augment data to achieve stronger text classification performance \cite{ye2022zerogen,yu2023regen}. LLMs has also demonstrated a strong capability in generating high-quality reasoning steps \cite{hsieh2023distilling,wei2022chain,kojima2022large}. In this study we aim to distill the reasoning ability of LLMs to smaller models,
and encourage the models to distinguish the reasonablity of explanations to identify informative data in AL scenario. 

\textbf{Explanation Information}, as external knowledge, has been proven useful for a wide range of tasks in natural language processing \citep{hase2022models}. \citet{hase-etal-2020-leakage} used explanations as additional information and directly fed them into models. \citet{narang2020wt5} and \citet{Shen23} took the explanations as outputs and trained NLP models to generate them. 
How to leverage explanations is still an open problem \citep{hase2022models}.
In the active learning schema, some studies also attempt to leverage the explanations \citep{liang2020alice,wang2021teaching}, but they mainly focus on promoting the generalization abilities of models trained on low-resource data.
These AL studies are also hard to implement in text classification tasks and unlike these studies, we explore how to leverage explanations to identify informative unlabeled data for annotation.

%% file: method.tex
\section{Method}
\subsection{Overview}

\paragraph{Task Formulation} We mainly consider a $C$ class text classification task defined on a compact set $\mathcal{X}$ and a label space $\mathcal{Y} = \{1,...,C\}$. The data points are sampled i.i.d over the space  $\mathcal{Z} = \mathcal{X} \times \mathcal{Y}$ as $\{\textbf{x}_i,y_i\} \sim p_z$, which can be divided into two sets  -- the labeled set $D_l$ and the unlabeled set $D_u$. At the beginning of an active learning algorithm, only a small number of data points are randomly selected into the labeled set 
 $D_l$  and we have only access to data points in $D_l$ for training the classification model. Then $L$ data from $D_u$ are selected for annotation and added to $D_l$ (removed from $D_u$ simultaneously) in $\mathcal{M}$ multiple rounds. 
\paragraph{Model Architecture} Following previous work \citep{devlin2018bert}, we adopt a pre-trained bi-directional encoder as the backbone for classification. In addition to the encoder, a corresponding uni-directional decoder is applied to generate and score the explanation for the label prediction. During training, we construct $k$ different explanations $\textbf{e}_r$, i.e., $\{\textbf{e}_r\}_i, r=0,...,k-1,$ for each example $\{\textbf{x}_i,y_i\}$, where $\textbf{e}_0$ is the reasonable explanation and $\{\textbf{e}_{r>0}\}$ are $k-1$ unreasonable explanations. We leave the construction process of explanations in Section \ref{GenerationExp} for further descriptions. Before that, we will first present the model training and data selection in Section \ref{Training} and Section \ref{Selection} respectively. The framework of XAL is shown in Figure \ref{model} and the workflow can be found in  Algorithm \ref{alg1}.

\subsection{Training}
\label{Training}
For each text input $\textbf{x}$ (we omit all the subscripts of $i$ for simplicity in this subsection), we first prepend it with a special token $[CLS]$ and then obtain the contextual representation by feeding it into the encoder. The contextual representation of the $j$th token is calculated as:
\begin{equation}
    \textbf{h}_j = Encoder([CLS] + \textbf{x})[j].
    \label{cont_rep}
\end{equation}
The representation for [CLS], i.e., $\textbf{h}_0$ is taken as the sentence representation and fed into the classification layer, which is composed of a linear layer and a softmax function. The probability distribution on label space $\mathcal{Y}$ can be formulated as:
\begin{equation}
    P(y|\textbf{x}) = Softmax(Linear(\textbf{h}_0)).
\end{equation}
The cross-entropy loss is adopted to optimize the encoder parameters:
\begin{equation}
    \mathcal{L}_{cls} = - \sum P(y|\textbf{x}) \ log \ P({y}|\textbf{x}).
 \label{entropy}
\end{equation}
On the decoder side, the model is trained with teacher forcing to generate the golden explanation $\textbf{e}_{0}$. 
The generation loss is calculated as:
\begin{equation}
    \mathcal{L}_{gen} = - \sum _{t} log \ P (\textbf{e}_{0,t}|\textbf{h},\textbf{e}_{0,<t}).
\end{equation}
To make the decoder a good scorer to rank the reasonable and unreasonable explanations, we additionally adopt a ranking loss to optimize the decoder. In particular, the model is trained to rank between reasonable and unreasonable explanations. The ranking loss can be formulated as:
\begin{equation}
    \mathcal{L}_{rank} = \sum _{r>0} max(0, p_r - p_0),
\end{equation}
where $p_r$ is the explanation score for $\textbf{e}_r$,  calculated as the length-normalized conditional log probability:
\begin{equation}
    p_r = \frac{\sum _t log {P (\textbf{e}_{r,t}|\textbf{x},\textbf{e}_{r,<t})}}{||\textbf{e}_r||}.\label{score}
\end{equation}
The hyper-parameters are adopted to balance the weights of each loss, and the overall loss is formalized as follows:
\begin{equation}
    \mathcal{L} = \mathcal{L}_{cls} + \lambda _1 \mathcal{L}_{gen} + \lambda _2 \mathcal{L} _{rank}.
\end{equation}

\subsection{Data Selection in AL}
\label{Selection}
After training the model in each iteration, we can obtain an intermediate model $\pi$. To select the informative data in the unlabeled set $\mathcal{D}_u$, we adopt a combination of the predictive entropy and explanation score.  Specifically, for each raw data $\textbf{x}_i\in \mathcal{D}_u$, we first generate the explanation $\textbf{e}_{i}$ by selecting the top-$1$ output in the beam search. Then, we calculate the explanation score $p_{i}$ as Eq. \ref{score} and the predictive entropy $c_{i}$ as Eq. \ref{entropy}. The final score $s_i$ for example $\textbf{x}_i$ is calculated as the weighted sum of the normalized explanation score and predictive entropy:
\begin{equation}
    s_i =\frac{\lambda}{1+\lambda} \frac{e^{-p_i}}{\sum _i e^{-p_i}} + \frac{1}{1+\lambda}\frac{e^{c_i}}{\sum _i e^{c_i}}
    \label{rankingscore}
\end{equation}
where the $\lambda$ is the hyper-parameter to balance the explanation score and the predictive entropy. With the final score for each example, we rank the whole unlabeled instances and select the top $L$ instances for annotation.

\subsection{Generation of Golden Explanations}
\label{GenerationExp}
Previous work has shown that LLMs are good at reasoning \citep{bang2023multitask,rajasekharan2023reliable,hsieh2023distilling}. Inspired by these studies, we take the LLMs, such as ChatGPT and GPT4, as the teacher models, and query them to generate explanations for each selected labeled data, eliminating the annotation cost of human labor. In particular, we design slightly different prompt templates for different tasks, and the prompt for each task is shown in Appendix \ref{TASKANDPROMPTS}.  Taking stance detection as an example, its prompt template is designed as \textit{`The stance of this tweet to the target \{\textbf{Target}\} is \{\textbf{Label}\}, explain the reason within 50 words’}, where the \textbf{Target} is the corresponding stance target, and the \textbf{Label} is the classification label. The final query to the teacher model is the concatenation of the text and the prompt. We construct a reasonable explanation by feeding the golden label into the query and generate several unreasonable explanations by feeding wrong labels. Figure \ref{explanation} shows an example that we generate explanations by querying ChatGPT, where we can observe that ChatGPT could provide different explanations according to the given label.

%% file: exp.tex
\section{Experiments}

\begin{table*}[] \small 
 \centering
\begin{tabular}{lp{4cm}cccc}
\hline
 Task & Dataset & \# Labels & Train& Dev& Test  \\ \hline
Natural Language Inference & RTE \citep{bentivogli2009fifth} &2 &2,240 &250 &278\\
Paraphrase Detection & MRPC \citep{dolanunsupervised}&2 &3,667&409 &1,726  \\
Stance Detection & COVID19 \citep{glandtstance}&3 & 4,533& 800&800 \\
 Category Sentiment Classification & MAMS \citep{jiang2019challenge}  &3 &7,090 & 888&901\\
(Dis)agreement Detection & DEBA \citep{pougue2021debagreement} &3 & 4,617 & 578&580\\
Relevance Classification & CLEF \citep{Kanoulas}&2 & 7,847&981&982\\
\hline 
\end{tabular}   
\caption{All the six text classification tasks used in our experiments. The extent of difficulty is roughly in an increasing tendency.  RTE and MRPC are fundamental natural language tasks and are included in the widely used benchmark GLUE. MAMS, COVID19, and DEBA require the model to understand the text and give suitable inferences towards a specific target or text, and CLEF further provides a difficult dataset with imbalanced label distribution.
} 
\label{tasks}
\vspace{-2mm}
\end{table*}

\subsection{Tasks and Dataset}
We conduct experiments on six different text classification tasks: (1) \textbf{Natural Language Inference} aims to detect whether the meaning of one text is entailed (can be inferred) from the other text; (2) \textbf{Paraphrase Detection} requires identifying whether each sequence pair is paraphrased; (3) \textbf{Category Sentiment Classification} aims to identify the sentiment (Positive/Negative/Neutral) of a given review to a category of the target such as food and staff; (4) \textbf{Stance Detection} aims to identify the stance (Favor/Against/Neutral) of a given text to a target; (5) \textbf{(Dis)agreement Detection} aims to detect the stance (Agree/Disagree/Neutral) of one reply to a comment; (6) \textbf{Relevance Classification} aims to detect whether a scientific document is relevant to a given topic. The details of the dataset we used are shown in Table \ref{tasks}.
Appendix \ref{TASKANDPROMPTS} demonstrates the details and prompts of six datasets with examples. \footnote{Without losing generality, we randomly split the training set in RTE, and MRPC into train/dev set with proportion 9:1. In DEBA, we adopt the topic of climate change for experiments.}


\subsection{Baselines}
To demonstrate the effectiveness of our proposed method, we compare XAL with the following nine AL baselines:
(1) \textbf{Random}  uniformly selects unlabeled data for annotation; (2) \textbf{Max-Entropy (ME)} \citep{lewis1995sequential,schohn2000less} calculates the predictive entropy in the current model and selects data with max entropy ;
(3) \textbf{Bayesian  Active Learning by Disagreement (BALD)} \citep{houlsby2011bayesian} exploits the uncertainty of unlabeled
data by applying different dropouts at test time;
(4) \textbf{Breaking Ties (BK)} \citep{scheffer2001active} selects instances with the minimum margin between the top two most likely probabilities ;
(5) \textbf{Least Confidence (LC)} \citep{culotta2005reducing} adopts instances whose most likely label has
the least predictive confidence; 
(6) \textbf{Coreset} \citep{sener2018active,ChaiCore} treats the representations in $D_u$ as cluster centers, and selects the unlabeled data with the most significant distance from its nearest centers;
(7) \textbf{Batch Active learning by Diverse Gradient Embeddings  (BADGE)} \cite{ash2019deep} measures uncertainty as the gradient magnitude and collects examples where these gradients span a diverse set of directions; (8) \textbf{Bayesian Estimate of Mean Proper Scores (BEMPS)} \cite{tan2021diversity} encourages diversity in the vector of expected changes in scores for unlabelled data;  
(9) \textbf{Contrastive Active Learning (CAL)} \citep{margatina2021active} selects instances with the maximum
mean Kullback-Leibler (KL) divergence between its $m$ nearest neighbors.

 \begin{figure*}
    \centering
    \includegraphics[width=\hsize]{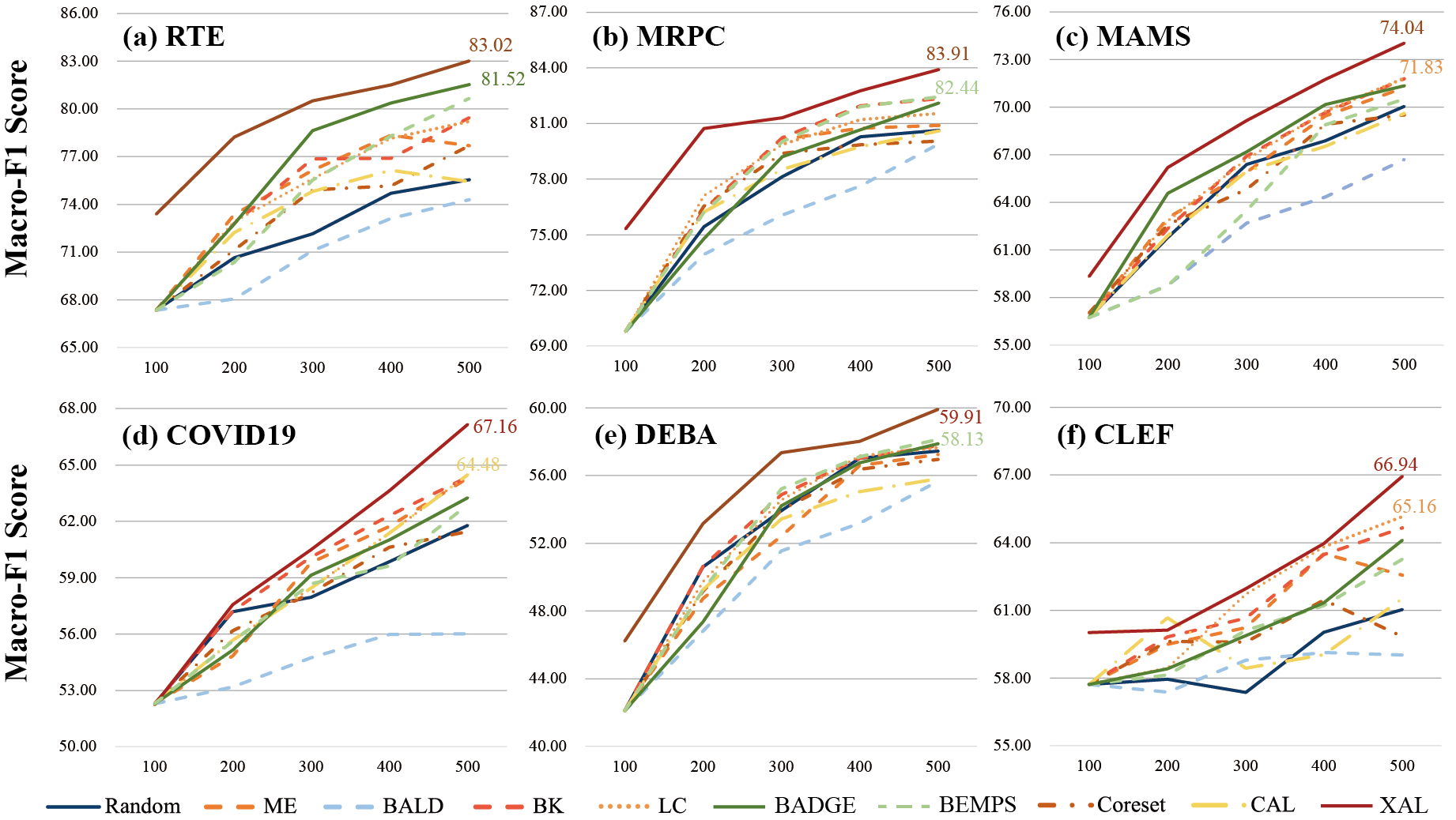}
    \vspace{-3mm}
    \caption{Results given the data selection budget 500 instances in six text classification tasks, where 100 instances are selected for annotation in each iteration.
    Here we plot the specific values of XAL and the second significant performance when using 500 instances, and the detailed performance values can be found in Appendix \ref{MainDetails}.}
    \label{MainDetailsFigure}
    \vspace{-2mm}
\end{figure*}

%% file: results.tex
\vspace{-1mm}
\section{Results and Discussion}
\label{Results}
\begin{figure*}[t]
    \centering
    \includegraphics[width=0.98\hsize]{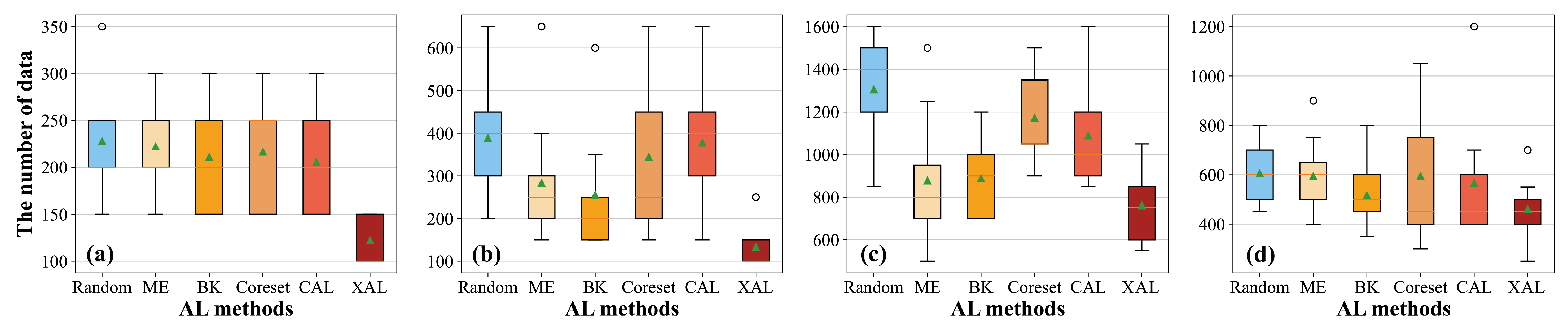}
    \vspace{-1mm}
    \caption{Experimental results demonstrate how much data, when selected using AL methods, is required for the models to achieve 90\% of the performance of those trained on the complete training datasets. In each iteration, we annotate 50 instances. The performance of models trained on the whole training sets is, (a)  RTE -- 83.11\%, (b) MRPC -- 84.74\%, (c) COVID19 -- 75.45\%, and (d) DEBA -- 65.71\%. The green triangles refer to the average values of the experiments on three different initial sets $D_l$ and three different random seeds. The circles refer to outliers. Detailed results can be seen in Appendix \ref{UPPERBOUND}. }
    \label{boxplot}
    \vspace{-2mm}
\end{figure*}

\subsection{Main Results}
\label{MainResults}
We mainly consider two different settings: (1) Given the data selection budget, we observe the 
trend of changes in model performance; (2) Given the performance upper bound, we observe the number of required instances that the model needs to achieve 90\% of the upper-bound performance. We utilize FLAN-T5-large \cite{chung2022scaling} as our backbone network \footnote{ We implement baselines using  FLAN-T5-large as well.}, ChatGPT is adopted to generate the explanations. The implemented details can be found in Appendix \ref{Implementation} and the detailed values of the result can be found in Appendix \ref{MainDetails}.  

\paragraph{Given Data Selection Budget} 
\label{Budget}
Following previous work \citep{zhang2017active,schroder-etal-2022-revisiting}, we set the data selection budget as 500 instances and select 100 instances for annotation in each iteration. The results are presented in Figure \ref{MainDetailsFigure}. 
We can observe that the proposed XAL model consistently outperforms other active learning methods. For instance, in MAMS, our model attains a macro-F1 score of 74.04\% at the end, which is 2.21\% higher than the second-best result (LC at 71.83\%). Similarly, in DEBA, XAL surpasses the second-best result (BADGE at 58.13\%) by 1.78\%. These results demonstrate the effectiveness of our XAL framework in addressing text classification tasks. 

In COVID19, while the model does not significantly outperform the baselines at the beginning (possibly due to the relatively high complexity of the task), it still exhibits stronger performance with a data count of 300-500, which underscores the effectiveness of the data selection in XAL. In CLEF, we notice that the performance of baseline models is notably unstable due to the significant imbalance in label distribution (the ratio between relevant and irrelevant is approximately 1:21). However, our XAL model achieves superior performance and more consistent improvements over baselines during the data selection process, which validates the effectiveness of XAL, even in challenging scenarios of imbalanced data. 

\paragraph{Given Performance Upper Bound} It's also valuable to evaluate the amount of data required for models to achieve comparable performance with those trained on the entire training dataset. Specifically, we begin with an initial labeled set of 100 instances and select a certain number of instances to annotate in each selection iteration \footnote{To balance the training efficiency and the performance gap, we set the selection number as 50.}, and cease the AL process once the model performance reaches 90\% of the upper-bound performance. Experimental results are depicted in Figure \ref{boxplot}.\footnote{For ease of presentation and training efficiency, we only report results on four tasks.} As observed, XAL requires the least amount of data to reach the performance goal. For instance, in the  task of DEBA, XAL necessitates an average of 461.11 data points, which is 55.56 less than the second lowest value (BK--516.67). To conclude, XAL models only require 6\%, 3\%, 16\%, and 10\% of the data from RTE, MRPC, COVID19, and DEBA tasks respectively to achieve 90\% performance of models that are trained on the entire datasets, which significantly reduces the annotation cost. These results show that the proposed XAL is very cost-efficient in selecting informative unlabeled data.

\begin{figure*}
    \centering
    \includegraphics[width=0.97\hsize]{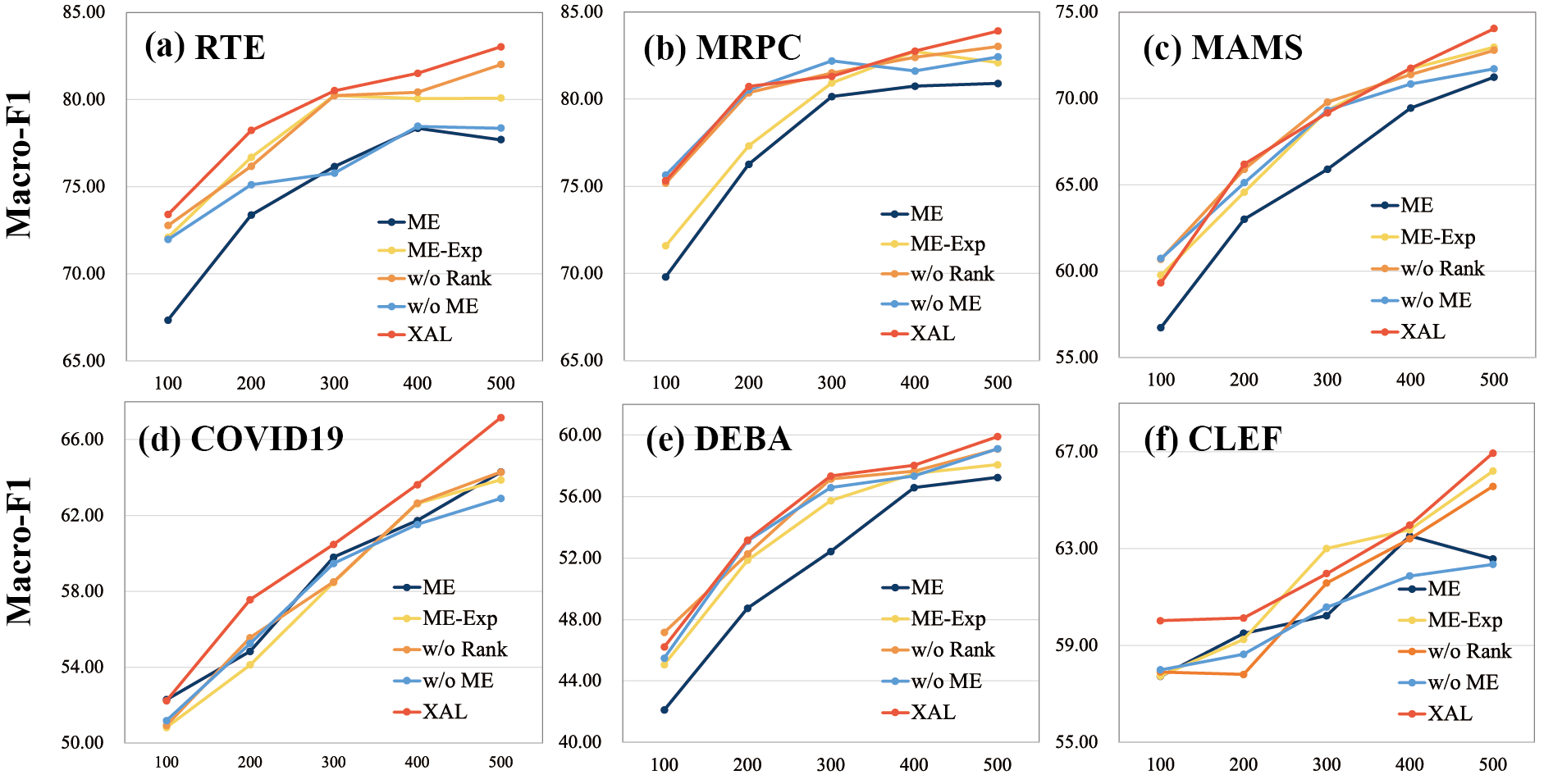}
    \vspace{-1mm}
    \caption{Results of ablation study in the six text classification tasks. We select 100 instances in each iteration and conduct 4 iterations (the same with Section \ref{Budget}). The results are measured using macro-F1 scores and they are the average values on three different initial sets $D_l$ and three different random seeds.}
    \label{abl}
    \vspace{-3mm}
\end{figure*}

\subsection{Ablation Study}
\label{Abalation}
We conduct an ablation study to investigate the impact of each module in our model. The results are displayed in Figure \ref{abl}.
Firstly, we conduct a comparison among ME, ME-Exp, and XAL, where ME-Exp has the same model structure as XAL but it selects the unlabeled data with the predicted classification entropy. We observe that ME-Exp can achieve superior performance on most datasets compared to ME, which demonstrates the effectiveness of using explanations. However, XAL further achieves noticeably better performance over ME-Exp, indicating that the improvement in XAL comes not only from the introduction of explanations but also from the data selection strategy (with explanation scores). Next, we compare XAL with a version that removes the ranking loss (\textit{w/o Rank} in Figure \ref{abl}). XAL also achieves better performance on most datasets and with different numbers of labeled data, indicating that the ranking loss can enhance the effectiveness of data selection in the AL process. Furthermore, the performance of selecting data solely using the explanation score but without using predictive entropy is also illustrated in Figure \ref{abl} (\textit{w/o ME}). We observe that removing ME leads to significant performance drops on most datasets, implying that the predictive entropy and explanation score can complement each other. 

\begin{table}[t]\small
\centering
\begin{tabular}{lccccc}
\hline
    & 100&	200&	300	&400&	500         \\
    \hline
ChatGPT & \textbf{60.79} &63.47 &68.51 &71.54 &73.24 \\
ALPACA-7B &59.52&61.75&67.77&71.12&72.24\\
GPT4 &59.67&\textbf{64.28}&\textbf{69.51}&\textbf{72.96} & \textbf{74.63}\\

\hline
\end{tabular}
\vspace{-2mm}
\caption{Model performance on MAMS using different explanation generations. We compare the performance in a certain initial set and random seed.}
\label{Generation}
\vspace{-1mm}
\end{table}

\begin{table}[t]\small
\centering
\begin{tabular}{lcccc}
\hline
   &	MRPC	&COVID19&DEBA &CLEF        \\
    \hline
Zero-shot &72.46&66.67&48.96&34.21\\
Few-shot &78.32&\textbf{67.73}&54.69&42.44\\
Silver &74.32&53.66&47.12&36.09\\
XAL (500) &\textbf{83.91}&67.16&\textbf{59.91}&\textbf{66.94} \\
\hline
\end{tabular} 
\vspace{-2mm}
\caption{Performance of ChatGPT and performance of models trained on `silver' data.}
\label{ChatGPT}
\vspace{-3mm}
\end{table}

To further evaluate how the ranking loss works in XAL, we also compare the model's capability to rank explanations between XAL and its counterpart without ranking loss. Experimental results show that XAL achieves superior performance. For instance, the ranking accuracy in RTE and MRPC for XAL are 73.93\% and 78.62\%, which are 5.36\% and 4.30\% higher than those without ranking loss, respectively (detailed results are shown in Appendix \ref{ScoreAPP}). These results suggest that the ranking loss can enhance the model's ability to score the explanations.  It is evident that XAL consistently outperforms these alternatives in most time, while there are some fluctuations across different scenarios.



\subsection{Explanation Generation}
\label{ExpGeneAnalyze}

We also carry out experiments to analyze how the generation of explanations impacts model performance. Specifically, we replace ChatGPT with ALPACA-7B \citep{alpaca}, and GPT4 \footnote{\href{https://openai.com/gpt-4}{https://openai.com/gpt-4}} to generate explanations on the MAMS dataset. The results are presented in Table \ref{Generation}. We also observe that the ALPACA-7B can also provide useful explanations to some extent and enhance the model performance compared with ME through our framework.  This suggests that LLMs, when used as an assistant in XAL, can provide consistent explanation generation and enhance model performance.  The results also indicate that the model performance can be affected by the explanation generation model and it is applicable to use open-source LLMs. The results of human annotation are also discussed in Appendix \ref{HumanAnnot}. 

\subsection{Comparison with ChatGPT}
We assess ChatGPT on these datasets in both zero-shot and few-shot scenarios, and the outcomes are presented in Table \ref{ChatGPT}. The models, when subjected to supervised fine-tuning with 500 labeled data points in XAL, exhibit either notably improved or comparable performance to ChatGPT across these datasets. This suggests that our model can achieve satisfactory results with minimal cost.

Since XAL queries ChatGPT for generating explanations with extra cost (API calls) to implement AL, we also analyze if we could obtain a better model through querying ChatGPT for more labeled data, i.e. `silver data' (the labels are probably incorrect). In detail, we use the actively selected data with golden labels, and randomly selected data with `silver' labels from ChatGPT to train the model in each active iteration. For this process, ChatGPT is employed to annotate a random selection of data from $D_u$, which is three times the size of $D_l$—mirroring the frequency of queries made for explanations in XAL. We report the final model performance with 500 golden labels and 1500 silver labels in  Table \ref{ChatGPT}.  We observe that despite the increase in silver data obtained from ChatGPT, our model can still perform more significantly. This outcome is primarily attributed to the uncertain accuracy of ChatGPT's annotations. The lack of reliability in these pseudo labels suggests that merely increasing the quantity of labeled data, without ensuring its quality, may not be an effective strategy for improving model performance.

\subsection{Human Evaluation on Interpretability}
We evaluate our model's ability to explain its prediction by examining the consistency between the generated explanation and the classification label. Specifically, we randomly select 50 test instances and use the model trained on 500 instances (see Section \ref{MainResults}) to generate the labels and explanations. Then we ask humans to infer the classification labels based solely on the generated explanations. The consistency is measured by whether the human-inferred label equals the label predicted by the model. We report the consistency rate across all the test sets: MRPC-94\%, RTE-94\%, COVID19-96\%, DEBA-94\%, MAMS-94\%, CLEF-100\%. We find that the consistency rates on all six tasks exceed 94\%, which demonstrates that XAL explains its classification prediction very well. Case studies for the generated explanations and the predicted labels are presented in Appendix \ref{CaseStudy}.

\begin{figure}[t]
    \centering
    \includegraphics[width=0.9\hsize]{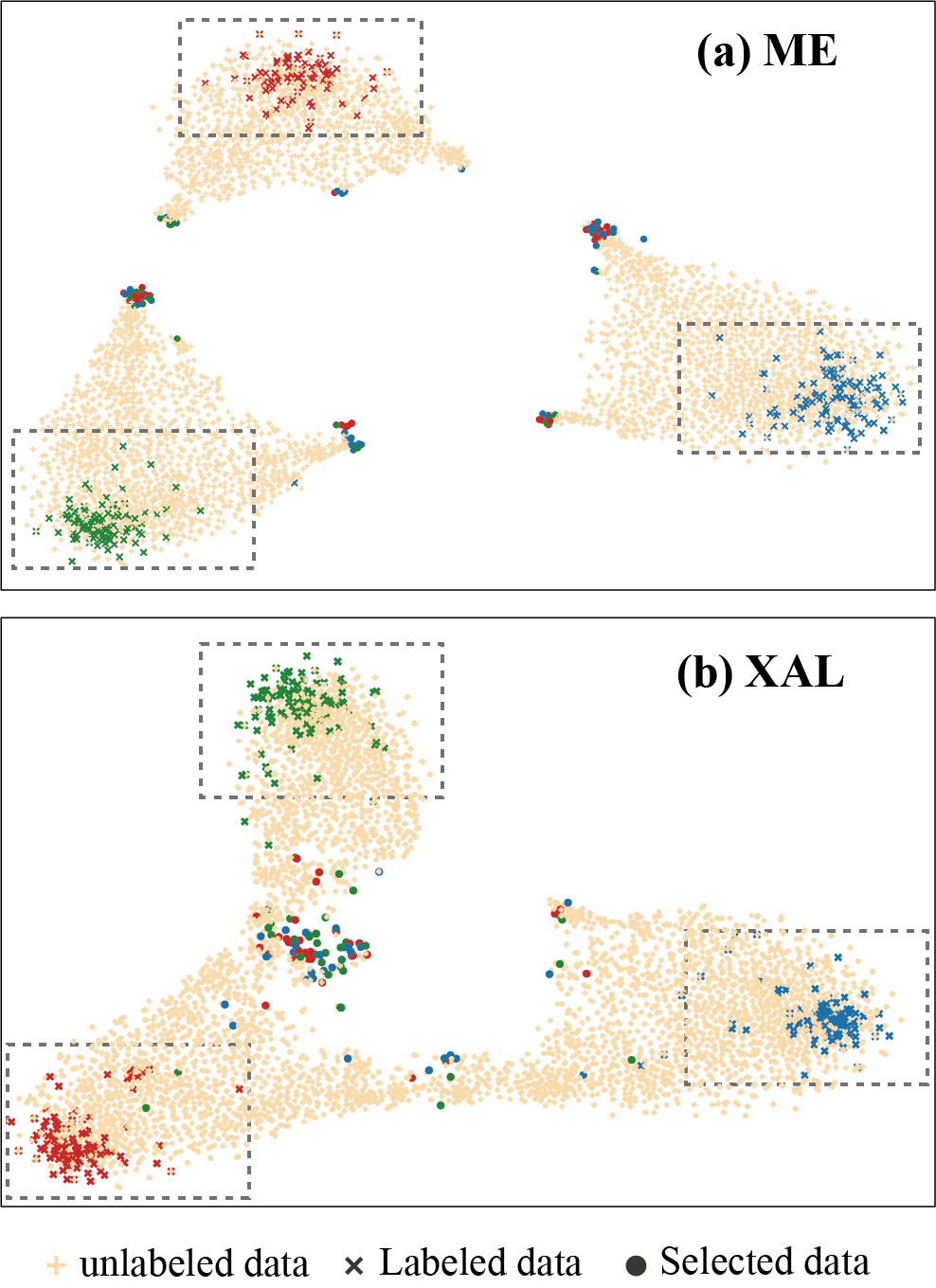}
    \vspace{-1mm}
    \caption{t-SNE visualizations of contextual representations. To facilitate identification, we outline the areas of labeled data with dashed squares. The colors, i.e. red, blue, and green, refer to the different labels.}
    \label{visulize}
    \vspace{-2mm}
\end{figure}  
\subsection{Representation Visualization}
\label{Visual}
To understand the potential of XAL in exploring informative unlabeled data, we use t-SNE \citep{tsne} to ``visualize'' the data selection procedure of ME and XAL on the task DEBA. Specifically, with the intermediate model in Section \ref{MainResults} (trained with 200 labeled instances),  100 instances from the unlabeled set $D_u$ are then selected for annotation. Then, we feed all the labeled and unlabeled instances into the model and get their sentence representations ($\textbf{h}_0$ in Eq. \ref{cont_rep}). Finally, we apply the t-SNE toolkit to map these representations into a two-dimensional embedding space, which is shown in Figure \ref{visulize}. We can observe that the data are obviously partioned into different clusters, which shows the overconfidence problem in ME, and the unlabeled data selected by ME is only distributed around the decision boundary, showing that the model can only select the high-uncertainty data it believes. However, the proposed XAL can select more diverse data, some of which are wrongly classified by the current model. These results demonstrate that the data selection strategy in XAL can identify more informative data and mitigate the problem of overconfidence to some extent. More visualizations are shown in Appendix \ref{VisAPP}.

%% file: conclusion.tex
\vspace{-1mm}
\section{Conclusion}
In this paper, we proposed a novel Explainable Active Learning (XAL) framework for text classification. Experiments demonstrated that XAL achieves substantial improvements compared with previous AL methods. 
Further analysis indicated that the proposed method can generate corresponding explanations for its predictions. 

\section{Limitations}
\label{Limitation}

XAL takes an initial attempt towards integrating rationales information into active learning. While acknowledging that this approach may necessitate additional computational resources, this augmentation empowers the trained classifier to be both more explainable and more generalized, as the model can generate explanations for its predictions and obtain enhanced performance. Our model, which incorporates a decoder module to obtain the generation score, necessitates more time for data selection, which is detailed in Appendix \ref{Cost}, but during the inference, since we use an encoder-decoder models for training, we can directly use the encoder for inference if the explanation generation is not in need.
In our experiments, we evaluated our model's effectiveness across six classification tasks in a low-resource setting, but XAL can be used for other tasks with more label classes and industrial downstream applications. 


\section{Ethical Statement}
We honor the ACL Code of Ethics. No private data or non-public information was used in this work. All annotators have received labor fees corresponding to the amount of their annotated instances. 

\section{Acknowledgement}
We acknowledge financial support of the National Natural Science Foundation of China Key Program under Grant Number 62336006.

%% file: appendix.tex
\appendix
\clearpage
\section{Tasks and Corresponding Prompts}
\label{TASKANDPROMPTS}
\begin{figure*}
    \centering
    \includegraphics[width=\hsize]{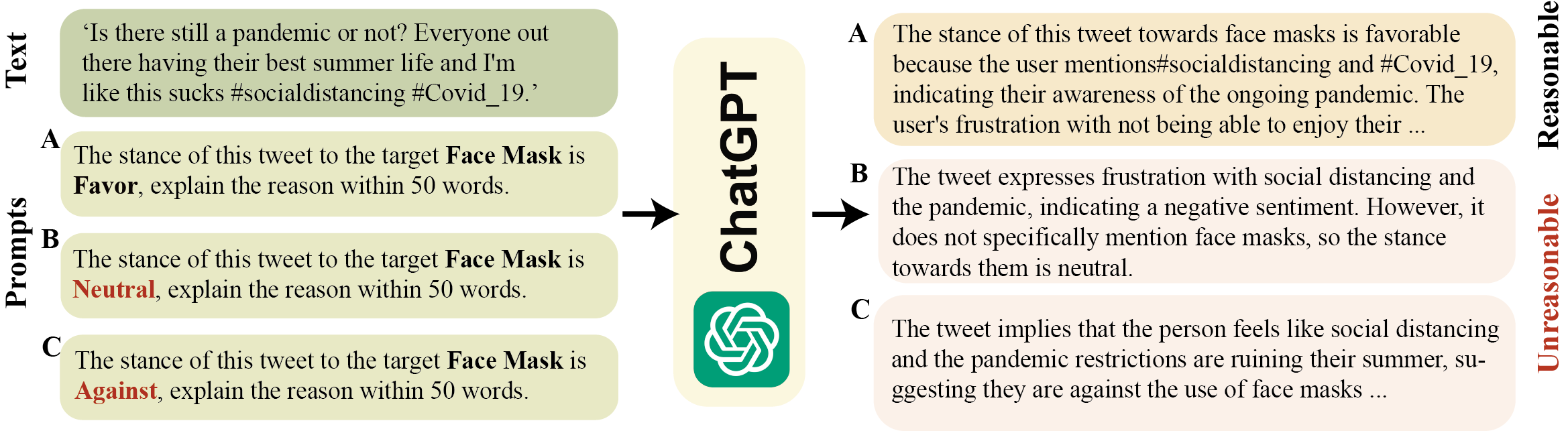}
    \caption{The process to generate diverse explanations from LLMs. We can obtain reasonable and unreasonable explanations by querying ChatGPT with correct and incorrect labels, respectively.}
    \label{explanation}
\end{figure*}

We show the tasks and examples for experiments in Table \ref{tasksandexamples}, including natural language inference, paraphrase detection, category sentiment classification, stance detection, (dis)agreement detection, and relevance classification. Then we show how we obtain different explanations in Figure \ref{explanation} with an example of COVID19. We also show the prompts we used for explanation generation through querying ChatGPT (Table \ref{tasksandprompst}).

\begin{table*}[h] \small
 \centering
\begin{tabular}{lp{9cm}}
\hline
 Task & Text   \\ \hline
Natural Language Inference & \textbf{Sentence 1}: Danny Kennedy, Greenpeace campaigns director, said: "The burden of proof in the Scott Parkin expulsion case lies morally with the Commonwealth, to prove that he is a danger. When the Government brought in anti-terror legislation, they promised the public that these laws would only be used to confront a real and present risk of a terrorist attack, not a sweep-all approach against citizens. Peace is not terrorism. Peace is not a threat to national security. No democratic government should expel a foreign citizen because [it] opposes his political opinions."

\textbf{Sentence 2}: Greenpeace director said that peace is terrorism.

\textbf{Label}: Not Entailment.
\\ \hline 
Paraphrase Detection & \textbf{Sentence 1}: Last week the power station’s  US owners, AES Corp, walked away from the plant after banks and bondholders refused to accept its financial restructuring offer .",

\textbf{Sentence 2}: "The news comes after Drax's American owner, AES Corp. AES.N, last week walked away from the plant after banks and bondholders refused to accept its restructuring offer.

\textbf{Label}: Paraphrase/Semantic Equivalent.
\\ \hline
Category Sentiment Classification &  \textbf{Text:} I left feeling unsatisfied, except for having a nice chance to people watch in the cozy atmosphere with my over-priced pasta bolognese.

\textbf{Target}: Ambience 

\textbf{Label}: Positive
\\ \hline
Stance Detection &  \textbf{Text:} Michigan is fining individuals 500\$ for not wearing a mask in public. How do y'all feel about this? Curious because I am torn about being so forceful but agree that people should wear masks. \#MaskOn.

\textbf{Target}: Face Mask

\textbf{Label}: Favor
\\ \hline
(Dis)agreement Detection & \textbf{Text 1}: True, but with lower power usage, you have less heat to dissipate, meaning you can overclock it even more.

\textbf{Text 2}: AMD creates a chip that saves energy by over 31 times. Someone show this to r/PCMasterRace cause we need to switch to AMD.

\textbf{Label}: Agree.
\\ \hline

Relevance Classification &\textbf{Document:} 99mtechnetium penicillamine: a renal cortical scanning agent. 99mTechnetium penicillamine, a renal cortical imaging agent, can be used to provide a rapid, safe, and non-invasive assessment of renal morphology and the renal vascular supply. Since this agent is not excreted significantly during the imaging procedure cortical scans of high quality can be obtained without image deterioration owing to a superimposed collecting system. These scans, which are clearly superior in anatomical detail to earlier scans using 131I hippuran, can be obtained along with the 131I hippuran renogram when the patient comes to the nuclear medicine department. Herein we demonstrate the anatomical detail it is now possible to achieve by presenting the cortical renal scans and accompanying radiograms from 5 patients with different renal pathology.

\textbf{Topic}: Procalcitonin, C-reactive protein, and erythrocyte sedimentation rate for the diagnosis of acute pyelonephritis in children.

\textbf{Label}: Not Relevant. \\

\hline 
\end{tabular}   
\caption{Tasks and examples for experiments.} 
\label{tasksandexamples}
\end{table*}


\begin{table*}[h] \small
 \centering
\begin{tabular}{p{2.5cm}p{7cm}p{3.5cm}}
\hline
 Task & Prompts &Label set \\ \hline
Natural Language Inference & Sentence 1: \textbf{\{Text 1\}}.  Sentence 2: \textbf{\{Text 2\}}.  Sentence 1 can \textbf{\{Label\}} sentence 2, explain the reason within 50 words. & \{Entail, Not Entail\} \\
Paraphrase Detection & Sentence 1: \textbf{\{Text 1\}}.  Sentence 2: \textbf{\{Text 2\}}.   The relation between the above two sentences is \textbf{\{Label\}}, explain the reason within 50 words. & \{Paraphrase Equivalent, Not Paraphrase Equivalent\} \\
Category Sentiment Classification &  \textbf{\{Text\}}. &\{Positive, Negative, Neutral\}\\
Stance Detection &  \textbf{\{Text\}}. The stance of this tweet to the target \textbf{\{Target\}} is \textbf{\{Label\}}, explain the reason within 50 words. &\{Favor, Neutral, Against\}\\
(Dis)agreement Detection & Sentence 1: \textbf{\{Text 1\}}.  Sentence 2: \textbf{\{Text 2\}}.   The stance of sentence 2 is \textbf{\{Label\}} to sentence 1, explain the reason within 50 words. & \{Agree, Neutral, Disagree\} \\
Relevance Classification & \textbf{\{Document\}}. The scientific document is \textbf{\{Label\}}' to the research topic \textbf{\{Topic\}}, explain the reason within 100 words. & \{Relevant, Not Relevant\} \\

\hline 
\end{tabular}   
\caption{The prompts we adopted for the tasks in our experiments. } 
\label{tasksandprompst}
\end{table*}

\section{Explanation Examples}
\label{ExpExample}
 Using the prompts in Appendix \ref{TASKANDPROMPTS}, we show some examples of the obtained explanations in Table \ref{Examples} by querying ChatGPT.

\begin{table*}[t] \small
 \centering
\begin{tabular}{p{5.8cm}p{1.2cm}p{5.8cm}}
\hline
Input &Label & Explanation \\ \hline
\textit{\textbf{RTE}}

\textbf{Sentence 1}: The main institutionalised forms of recognition for those who have made a significant contribution in the fields of physics, chemistry, medicine, literature, as well as for those working for peace (and more recently in the area of economics), are the Nobel prizes.

\textbf{Sentence 2}: Nobel Peace Prize candidates have been chosen. &\textbf{Not Entailment} & Sentence 1 states that the Nobel prizes are the main forms of recognition for various fields, including peace. However, sentence 2 only mentions that candidates have been chosen for the Nobel Peace Prize, without specifying if they have actually received the prize. Therefore, sentence 2 does not necessarily follow from sentence 1. \\
&Entailment & Sentence 1 can entail sentence 2 because sentence 1 mentions that one of the main forms of recognition for those working for peace is the Nobel prizes. Therefore, it is reasonable to assume that candidates for the Nobel Peace Prize have been chosen. \\ \hline
\textit{\textbf{MRPC}} 

\textbf{Sentence 1}: Maddox, 87, cracked two ribs when he fell about 10 days ago at an assisted living home where he was recovering from intestinal surgery, Virginia Carnes said.

\textbf{Sentence 2}: Maddox, who had battled cancer since 1983, cracked two ribs earlier this month when he fell at an assisted living home where he was recovering from surgery. & \textbf{ Equivalent} & The two sentences provide similar information but are expressed differently. Sentence 1 mentions that Maddox fell about 10 days ago, while Sentence 2 specifies that the fall occurred earlier this month. Despite the slight difference in wording, both sentences convey the same meaning. \\
&Not Equivalent &
The reason why the two sentences are not paraphrases or semantic equivalents is because they provide different information about the timing and reason for Maddox's rib injury. In the first sentence, it mentions that Maddox fell about 10 days ago while recovering from intestinal surgery, whereas the second sentence states that he fell earlier this month while recovering from surgery without specifying the exact time frame.\\
\hline
\textbf{\textit{COVID19}}

\textbf{Tweet:} Placed2nd @AngelaBelcamino Omg people say I have to wear a mask and I'm crying because I have to wear a mask to save lives boo hoo!! MAGAts are stupid!!!&\textbf{Favor} & The stance of this tweet towards face masks is favorable. The reason for this is that the tweet expresses support for wearing masks to save lives and criticizes those who oppose wearing masks, referring to them as ""MAGAts"" (a term often used to describe supporters of former President Donald Trump)." \\
&Against & The reason behind the stance against face masks in this tweet seems to be frustration and ridicule towards those who oppose wearing masks. The tweet suggests that the person finds the idea of wearing masks to save lives unnecessary and mocks those who support it, particularly referring to them as ""MAGAts,"" a derogatory term for supporters of former President Trump.  \\

&Neutral & The tweet expresses a neutral stance towards face masks by mocking the idea of crying over having to wear one. While it criticizes MAGA supporters, it does not explicitly state a negative or positive opinion about face masks themselves."
\\
\hline

\end{tabular}   
\vspace{-2mm}
\caption{Diverse explanations generated from ChatGPT. The golden label is marked in \textbf{bold}. } 
\vspace{-3mm}
\label{Examples}
\end{table*}

\section{Algorithm}
\label{Algorithm}
We show the detailed algorithm of XAL in Algorithm \ref{algrithm}.

\begin{algorithm} [h]
	\renewcommand{\algorithmicrequire}{\textbf{Input:}}
	\renewcommand{\algorithmicensure}{\textbf{Output:}}
	\caption{Explainable Active Learning Algorithm}
	\label{alg1}
	\begin{algorithmic}[1]
		\STATE Initialization: dataset $D_u$, iteration steps $\mathcal{M}$, selective number $L$, training epoch $\mathcal{T}$. 
            \STATE Randomly select $L$ data from $D_u$, denoted as $D_s$ and remove them in $D_u$.
            \STATE Annotate the data $\textbf{x}_i \in D_s$ for $y_i^c$ with human annotators.
            \STATE Query ChatGPT for diverse explanations $y_i^{g_r}$ for the data $\{\textbf{x}_i, y_i^c \} \in D_s$. 
            \STATE Add $\{\textbf{x}_i,y_i^c, y_i^{g_r}\} \in D_s$ to $D_l$, and empty the set $D_s$. 
            \STATE $m=1$. 
		\REPEAT
		\STATE $m \leftarrow m + 1$
            \STATE Initialize an explainable classifier $\pi$ and $t = 0$.
            \REPEAT 
            \STATE $t \leftarrow t + 1$
            \STATE Calculate optimization loss using data $\{\textbf{x}_i,y_i^c, y_i^{g_r}\} \in D_l$.
            \STATE Optimize the explainable classifier $\pi$.
            \UNTIL $t>\mathcal{T}$
            \STATE Calculate the predictive entropy $\textbf{p}_i$ and explanation scores $\textbf{c}_i$ of data $\textbf{x}_i \in D_u$ uing Eq. \ref{score}.
            \STATE  Calculate the rank score using Eq. \ref{rankingscore}.
		\STATE Select $L$ data with the largest score from $D_u$ to  $D_s$.
            \STATE Annotate the data in $D_u$ following the steps 3-5.
		\UNTIL $m>\mathcal{M}$
		\ENSURE  Explainable classifier $\pi$.
	\end{algorithmic}  
 \label{algrithm}
\end{algorithm}


\section{Implementation Details} 
\label{Implementation}

In our experiments, we directly utilize a pre-trained encoder-decoder language model for its strong ability in text understanding and generation. Specifically, we adopt the officially released pre-trained FLAN-T5-Large model \citep{chung2022scaling} from Huggingface \footnote{\href{https://huggingface.co/}{https://huggingface.co/}}.
All models in our experiments are trained on a single GPU (Tesla V100) using the Adam optimizer \citep{kingma2014adam}. We set the learning rate at 1e-4, with a linear scheduler. The batch size is consistently set to 1 across all tasks. 
The models are trained for 10 epochs in each iteration. Hyper-parameters $\lambda _1$, $\lambda _2$, and $\lambda$ are set to 0.1, 0.01, and 0.5, respectively, based on preliminary experiments.  Note that here we don not specially tune the hyperparameters using grid search, but select the parameters considering their magnitude and keep the same across different tasks, and we further implement sensitivity analysis in Appendix \ref{Sensitivity}. 
The performance for all tasks is evaluated based on macro-averaged F1. The reported results are the average of three initial sets $D_l$ and three random seeds (the average of 9 experimental results overall).

\section{Detailed Results}
\label{Detailed}
\subsection{Main Results}
\label{MainDetails}

The details of the main results are shown in Table \ref{mainresultstable}.

\begin{table*}[h]\small
\centering
\begin{tabular}{ccccccccccc}
\hline
    & Random & ME    & BALD  & BK             & LC    & Coreset & CAL          &BADGE &BMEPS  & XAL           \\
    \hline
    \textbf{\textit{RTE}} \\
    \hline
100 & 67.34  & 67.34 & 67.34 & 67.34          & 67.34 & 67.34   & 67.34    &67.34 &67.34       & \textbf{73.40} \\
200 & 70.64  & 73.37 & 68.06 & {72.80} & 72.91 & 71.12   & {72.22} & 72.76&70.36& \textbf{78.22} \\
300 & 72.16  & 76.15 & 71.09 & 76.85          & 75.60 & 74.90   & 74.81    &78.64&75.53      & \textbf{80.51} \\
400 & 74.71  & 78.35 & 73.11 & 76.90          & 78.15 & 75.16   & 76.15   &80.37&78.29       & \textbf{81.50} \\
500 & 75.54  & 77.69 & 74.30 & 79.44          & 79.22 & 77.69   & 75.42  &81.52&80.66        & \textbf{83.02} \\
\hline
\\
\textbf{\textit{MRPC}} \\
\hline
100 & 69.80  & 69.80 & 69.80 & 69.80          & 69.80 & 69.80   & 69.80  & 69.80 &69.80        & \textbf{75.31} \\
200 & 75.44  & 76.26 & 73.95 & {76.35} & 77.10 & 76.54   & {76.22} & 74.78&76.21& \textbf{80.73} \\
300 & 78.12  & 80.14 & 76.07 & 80.23          & 79.87 & 79.39   & 78.52 &79.21&80.02         & \textbf{81.31} \\
400 & 80.28  & 80.74 & 77.64 & 81.95          & 81.21 & 79.85   & 79.76    &80.67&  81.91    & \textbf{82.76} \\
500 & 80.63  & 80.90 & 79.90 & 82.33          & 81.53 & 80.06   & 80.60   &82.11& 82.44      & \textbf{83.91} \\

\hline \\ \textbf{\textit{MAMS}} \\
\hline
100 & 56.73  & 56.73 & 56.73 & 56.73 & 56.73 & 56.73   & 56.73 & 56.73 &  56.73 & \textbf{59.32} \\
200 & 61.77  & 63.01 & 58.75 & 62.34 & 62.83 & 62.59   & 61.89 & 64.57 & 59.64&  \textbf{66.19} \\
300 & 66.38  & 65.90 & 62.68 & 66.92 & 66.72 & 64.83   & 65.96 & 67.18 &64.48& \textbf{69.16} \\
400 & 67.88  & 69.44 & 64.33 & 69.67 & 69.74 & 68.93   & 67.54 & 70.26& 69.88 & \textbf{71.74} \\
500 & 70.05  & 71.23 & 66.69 & 71.78 & 71.83 & 69.50   & 69.59 & 71.35 & 70.51& \textbf{74.04} \\\hline

\\ \textbf{\textit{COVID19}} \\
\hline
100 & 52.29  & 52.29 & 52.29 & 52.29          & 52.29 & 52.29 & 52.29 & 52.29  & 52.29 & 52.24           \\
200 & 57.19  & 54.84 & 53.19 & {57.22} & 55.67 & 56.18   & 55.67 & 55.14 &	55.62&\textbf{57.57}           \\
300 & 57.95  & 59.80 & 54.74 & 60.10          & 58.45 & 58.18   & 58.45 &59.13&	58.67& \textbf{60.48} \\
400 & 59.85  & 61.73 & 55.98 & 62.30          & 61.38 & 60.62   & 61.38 & 61.01&	59.63&\textbf{63.63} \\
500 & 61.78  & 64.30 & 56.01 & 64.36          & 64.48 & 61.45   & 64.48 & 63.25	&62.88&\textbf{67.16} \\\hline

\\ \textbf{\textit{DEBA}} \\ \hline

100 & 42.09  & 42.09 & 42.09 & 42.09          & 42.09 & 42.09   & 42.09 &42.09 & 42.09 & \textbf{46.21} \\
200 & 50.60  & 48.74 & 46.81 & {50.65} & 49.73 & 49.18   & 49.26 & 47.35	&49.24&\textbf{53.16} \\
300 & 53.93  & 52.43 & 51.54 & 54.87          & 54.57 & 53.97   & 53.43 &54.21&	55.22& \textbf{57.35} \\
400 & 57.03  & 56.58 & 53.18 & 57.02          & 57.15 & 56.37   & 55.06 & 56.75&	57.12&\textbf{58.03} \\
500 & 57.45  & 57.25 & 55.66 & 57.78          & 57.64 & 56.95   & 55.82 &57.88	&58.13& \textbf{59.91} \\\hline

\\ \textbf{\textit{CLEF}} \\ \hline

100 & 57.72  & 57.72 & 57.72 & 57.72          & 57.72 & 57.72   & 57.72  & 57.72 & 57.72        & \textbf{60.02} \\
200 & 57.95  & 59.50 & 57.38 & {59.82} & 58.44 & 59.62   & \textbf{60.67} &58.42	&58.14 & {60.13} \\
300 & 57.37  & 60.23 & 58.80 & 60.66          & 61.72 & 59.60   & 58.44     &59.88&	60.12     & \textbf{61.97} \\
400 & 60.04  & 63.52 & 59.14 & 63.48          & 63.81 & 61.46   & 59.04    &61.34	&61.22     & \textbf{63.97} \\
500 & 61.04  & 62.57 & 59.03 & 64.66          & 65.16 & 59.84   & 61.53      &64.12&	63.27    & \textbf{66.94} \\
\hline
\end{tabular}
\caption{Main results in the six text classification tasks. We select 100 instances in each iteration and conduct 4 iterations. The results are measured using macro-F1 scores and they are the average values on three different initial sets $D_l$ and three different random seeds.}
\label{mainresultstable}
\end{table*}

\subsection{Given upper bound}
\label{UPPERBOUND}
We show the average number of data required for the model to achieve 90\% performance of those trained on all the training data (Table \ref{HowMuch}).
\begin{table*}[h]\small
\centering
\begin{tabular}{lcccccc}
\hline
    & Random&	ME&	BK	&Coreset&	CAL&	XAL         \\
    \hline
RTE     & 388.89 & 283.33 & {255.56} & 344.44    & {377.78} & \textbf{133.33} \\
MRPC    & 227.78 & 222.22 & 211.11          & 216.67    & {205.56} & \textbf{122.22} \\
COVID19 & 1305.56 &	877.78 &	888.89 	&1172.22 &	1088.89 &	\textbf{761.11}        \\
DEBA    & 605.56 & 594.44 & 516.67          & 594.44    & 566.67          & \textbf{461.11} \\
\hline
\end{tabular}
\caption{The detailed experimental results about how much data queried by AL methods can the model achieve 90\% performance of the models trained on the whole training data. In each iteration, we select 50 data. The model performances trained on the whole training sets are, (a)  RTE -- 83.11\%, (b) MRPC -- 84.74\%, (c) COVID19 -- 75.45\%, and (d) DEBA -- 65.71\%. The green triangles refer to the average values of the nine-times experiments.}
\label{HowMuch}
\end{table*}

\subsection{Ablation Study}
The detailed results of the ablation study are shown in Table \ref{DetailedAblation}.
\begin{table*}[h]\small
\centering
\begin{tabular}{cccccc}
\hline
    & ME & ME-Exp    & w/o Rank  & w/o ME                & XAL           \\
    \hline
    \textbf{\textit{RTE}} \\
    \hline
100 & 67.34 & 72.09 & 72.77 & 71.96 & \textbf{73.40} \\
200 & 73.37 & 76.68 & 76.17 & 75.10 & \textbf{78.22} \\
300 & 76.15 & 80.23 & 80.22 & 75.77 & \textbf{80.51} \\
400 & 78.35 & 80.05 & 80.42 & 78.46 & \textbf{81.50} \\
500 & 77.69 & 80.08 & 82.01 & 78.36 & \textbf{83.02} \\
\hline
\\
\textbf{\textit{MRPC}} \\
\hline 
100 & 69.80 & 71.58 & 75.18 & 75.64          & \textbf{75.31} \\
200 & 76.26 & 77.32 & 80.37 & 80.53          & \textbf{80.73} \\
300 & 80.14 & 80.93 & 81.50 & \textbf{82.19} & {81.31} \\
400 & 80.74 & 82.72 & 82.40 & 81.61          & \textbf{82.76} \\
500 & 80.90 & 82.09 & 83.02 & 82.42          & \textbf{83.91} \\
\hline
\\
\textbf{\textit{MAMS}} \\
\hline 
100 & 56.73 & 59.77 & \textbf{60.69} & 60.73          & {59.32} \\
200 & 63.01 & 64.57 & 65.90          & 65.11          & \textbf{66.19} \\
300 & 65.90 & 69.32 & \textbf{69.79} & {69.32} & {69.16} \\
400 & 69.44 & 71.71 & 71.38          & 70.83          & \textbf{71.74} \\
500 & 71.23 & 72.97 & 72.79          & 71.71          & \textbf{74.04} \\
\hline
\\
\textbf{\textit{COVID19}} \\
\hline 
100 & \textbf{52.29} & 50.83 & {50.94} & 51.19          & {52.24} \\
200 & 54.84 & 54.13 & 55.56          & 55.25          & \textbf{57.57} \\
300 & 59.80 & 58.48 & {58.51} & {59.48} & \textbf{60.48} \\
400 & 61.73 & 62.63 & 62.66          & 61.53          & \textbf{63.63} \\
500 & 64.30 & 63.88 & 64.29          & 62.90          & \textbf{67.16}
\\
\hline
\\
\textbf{\textit{DEBA}} \\
\hline 
100 & 42.11 & 45.06 & \textbf{47.16} & 45.48          & {46.21} \\
200 & 48.74 & 51.86 & 52.26          & 53.11          & \textbf{53.16} \\
300 & 52.43 & 55.74 & {57.15} & {56.59} & \textbf{57.35} \\
400 & 56.58 & 57.51 & 57.65          & 57.35          & \textbf{58.03} \\
500 & 57.25 & 58.08 & 59.09          & 59.11          & \textbf{59.91} 
\\
\hline 
\\
\textbf{\textit{CLEF}} \\
\hline
100 & 57.72 & 57.74 & {57.91} & 57.99          & \textbf{60.02} \\
200 & 59.50 & 59.24 & 57.80          & 58.63          & \textbf{60.13} \\
300 & 60.23 & 63.00 & {61.58} & {60.58} & \textbf{61.97} \\
400 & 63.52 & 63.78 & 63.40          & 61.87          & \textbf{63.97} \\
500 & 62.57 & 66.20 & 65.57          & 62.35          & \textbf{66.94}

\\
\hline

\end{tabular}
\caption{Detailed results of ablation study in the six text classification tasks. We select 100 instances in each iteration and conduct 4 iterations. The results are measured using macro-F1 scores and they are the average values on three different initial sets $D_l$ and three different random seeds.}
\label{DetailedAblation}
\end{table*}

\subsection{Capacity of Score}
\label{ScoreAPP}
To assess our model's capability to distinguish between reasonable and unreasonable explanations, we evaluate its ranking performance on the test set. Specifically, after four iterations of the AL process as per section \ref{MainResults}, we prompt ChatGPT to generate diverse explanations for the test data and score them using Eq. \ref{score}. In each test step, we feed both a reasonable and an unreasonable explanation to our model and calculate the accuracy in predicting the reasonable ones based on the computed explanation score (Table \ref{scoring}). As seen in the results, the model incorporating ranking loss achieves superior performance compared to the model without it. For instance, the accuracy in RTE and MRPC are 73.93\% and 78.62\% in the model with ranking loss, which are 5.36\% and 4.30\% higher than those without ranking loss, respectively. The improvement in prediction accuracy suggests that the ranking loss can enhance the model's ability to score the reasonability of explanations.
\begin{table*}[t]\small
\centering
\begin{tabular}{lcccccc}
\hline
    & RTE&	MRPC&	MAMS	&COVID19&	DEBA&	CLEF         \\
    \hline
w Ranking Loss &73.93&78.62&68.64&62.25&59.78&90.73\\
w/o Ranking Loss &68.57&74.32&62.04&58.34&55.04&87.39\\
\hline
\end{tabular}
\vspace{-1.5mm}
\caption{The performance of scoring the explanations w/o ranking loss.}
\label{scoring}
\vspace{-4mm}
\end{table*}

\section{Human Annotation}
\label{HumanAnnot}
We also carry out experiments to analyze how the human generation of explanations impacts model performance. Specifically, we replace ChatGPT with human annotation to generate explanations on the MAMS dataset.  For human annotation, three PhD students specializing in NLP annotate the labels and explanations. Specifically, the models achieve the  macro-F1 scores of RTE-62.13\%, MRPC-63.36\%, MAMS-67.38\%, COVID19-69.70\%, and CLEF-71.56\%, which are relatively lower compared to ChatGPT, which could be due to inconsistent annotation styles among annotators and changes in the annotation scheme from the original dataset \citep{gilardi2023chatgpt,zhu2023can}. The results also demonstrate the effectiveness of explanation generation through LLMs in XAL.

\section{Cost Analysis}
\label{Cost}

We conduct experiments (on one GPU V100 Tesla) to analyze the time consumption during each data query process in the MAMS task, which involves 7,090 training data instances. The results are as follows: ME-2 minutes, CA-2 minutes, BK-2 minutes, LC-2 minutes, BALD-11 minutes, Coreset-54 minutes, and our model XAL-21 minutes. Upon observation, it's apparent that our model requires more time for querying unlabeled data when compared to methods that leverage model uncertainty. However, it consumes less time than the representativeness-based method Coreset.


\section{Sensitivity Analysis}
\label{Sensitivity}
\begin{figure}
    \centering
    \includegraphics[width=0.95\hsize]{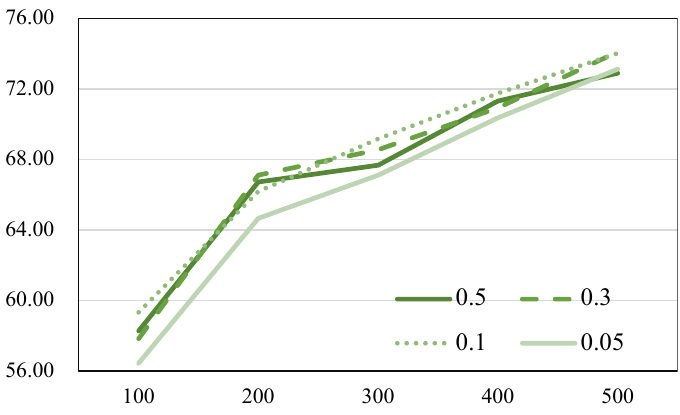}
    \caption{The experimental results with different hyper-parameter $\lambda_1$ in MAMS.}
    \label{lambda1}
\end{figure}

\begin{figure}
    \centering
    \includegraphics[width=0.97\hsize]{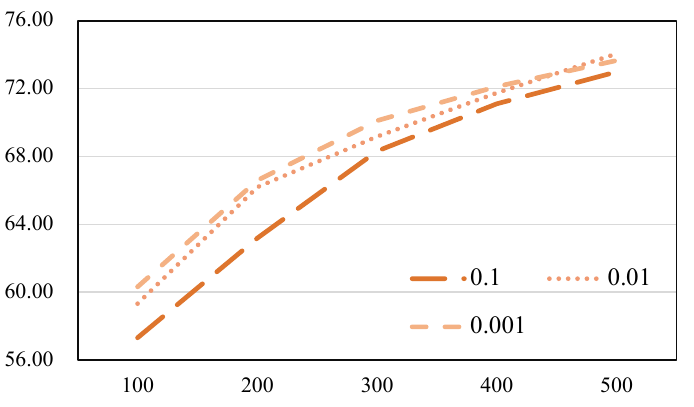}
    \caption{The experimental results with different hyper-parameter $\lambda_2$ in MAMS.}
    \label{lambda2}
\end{figure}

In this study, we establish our hyper-parameters based on the relative magnitude and importance of various loss functions, consistently applying these across all datasets without resorting to grid search for optimization. This section further explores the sensitivity of our model to these hyper-parameters.
Initially, we examine the impact of different $\lambda_1$ values, as depicted in Figure \ref{lambda1}. Our observations reveal that the model's performance remains relatively stable with $\lambda_1$ values of 0.3 and 0.1. However, a notable decline in performance occurs when $\lambda_1$ is increased to 0.5, attributed to the generative loss becoming approximately ten times greater than the classification loss. Conversely, reducing $\lambda_1$ to 0.05 results in a significant deterioration in model performance, suggesting that excessively minimizing the generative loss is detrimental.
Subsequently, we assess the model's response to various $\lambda_2$ values, detailed in Figure \ref{lambda2}. These findings indicate that higher $\lambda_2$ values can adversely affect the model's performance. Yet, the model exhibits lesser sensitivity to changes in $\lambda_2$ when it is equal to or less than 0.01.

\section{Case Study}
\label{CaseStudy}

\subsection{Model Generation}
Some generation cases are shown (Table \ref{CASE}) from the models trained for 500 data in the AL process of Section \ref{MainResults}. In these cases, we can find that our model can generate reasonable explanations for the label, which indicates the interpretability of our inference. But in some cases such as the case \romannumeral4., although the explanation generates the correct label Agree, it explains the label with a wrong reason, which implies that the explainer does not perform perfectly in the small number of data. But it also indicates that we can enhance the model performance in inference and generation by selecting the data with unreasonable explanations through human beings.

\begin{table*}[h] \small
 \centering
\begin{tabular}{p{6cm}p{1cm}p{6cm}}
\hline
Input &Golden & Explanation \\ \hline
\romannumeral1. \textit{\textbf{RTE}}

\textbf{Sentence 1}: Dana Reeve, the widow of the actor Christopher Reeve, has died of lung cancer at age 44, according to the Christopher Reeve Foundation.

\textbf{Sentence 2}: Christopher Reeve had an accident. & Not Entailment & Sentence 1 states that Dana Reeve, the widow of Christopher Reeve, has died of lung cancer at the age of 44. It does not provide any information about what caused Christopher Reeve's accident. Therefore, sentence 1 does not entail sentence 2.
\\
\hline
\romannumeral2. \textit{\textbf{MRPC}}

\textbf{Sentence 1}: Air Commodore Quaife said the Hornets remained on three-minute alert throughout the operation.

\textbf{Sentence 2}: Air Commodore John Quaife said the security operation was unprecedented.
& Not Equivalent
& The reason why the two sentences are not paraphrases or semantic equivalents is because they provide different information. Sentence 1 states that the Hornets were on three-minute alert during the operation, while Sentence 2 states that the security operation was unprecedented. 
\\

\hline
\romannumeral3. \textit{\textbf{MAMS}}

\textbf{Review}: Waiter took our drink order and then we didn't see him for 15 minutes.

\textbf{Category}: Stuff.  & Negative & The reason for the negative sentiment towards the staff in this review is because the waiter took the drink order and disappeared after 15 minutes. This indicates a lack of professionalism and attentiveness, which can be frustrating for customers.\\\hline

\romannumeral4. \textit{\textbf{COVID19}}

{\textbf{Tweet}}:
@WCheesemaker @StephenKing No. You’re missing the point. And from the way I phrased my statement I can understand why. I was attempting to point out masks work! They help stop the spread of disease. I should have stated it more clearly.
I’m a scientist; a big fan of masks- and science!! :relaxed: 

{\textbf{Target}}: Face mask.
& Favor & The stance of this tweet is in favor of face masks. The reason given is that the tweeter believes masks are effective in stopping the spread of disease, and therefore should be encouraged to use them. \\

\hline

\romannumeral5. \textit{\textbf{DEBA}}

{\textbf{Sentence 1}}: Best thing you can do is simply not drive. Take public transportation instead. 

{\textbf{Sentence 2}}: That's hugely unrealistic for many people living in places with poor public transport provisions (rural areas, poorer areas) or who live in poorly planned towns where amenities are spatially distant.
& Disagree & Sentence 2 disagrees with Sentence 1 because it argues that taking public transportation instead of driving is not feasible for many people living in places with poor public transport provisions or in poorly planned towns.
\\\hline
\romannumeral6. \textit{\textbf{DEBA}}

{\textbf{Sentence 1}}: I can't imagine believing Joe "Nothing Will Fundamentally Change" Biden is going to do anything except continue to serve his corporate masters.  This is...  uh, not a guy who's gonna accomplish much in office.

{\textbf{Sentence 2}}: Biden has surprised me with some of his action but on the whole it's still up to activists and local politicians to get stuff done.  Shame
& Agree
&
Sentence 2 agrees with Sentence 1 because it acknowledges that Biden has made some progress, but it also emphasizes that it is up to activists and local politicians to bring about meaningful change.

\\
\hline 
\end{tabular}   
\vspace{-2mm}
\caption{Case study of the explanation generation of our model. The model is trained on 500 labeled data instances following the AL process in Section \ref{MainResults}. } 
\vspace{-3mm}
\label{CASE}
\end{table*}

\subsection{Unreasonable Generation}
We also show some cases that our model believes have high unreasonability in the training set (Table \ref{BadCase}). It is noted that in these cases the model generates some unreasonable explanations. 

\begin{table*}[h] \small
 \centering
\begin{tabular}{p{6cm}p{1cm}p{6cm}}
\hline
Input &Golden & Explanation \\ \hline
\romannumeral1. \textit{\textbf{MAMS}}

\textbf{Review}: I like the smaller portion size for dinner.

\textbf{Category}: miscellaneous
& Positive
& The reason for the neutral sentiment polarity towards the aspect "miscellaneous" in the review is that the reviewer prefers the smaller portion size for dinner. This suggests that the overall dining experience is more enjoyable and satisfying, resulting in a neutral sentiment polarity. \\ 
\hline
\romannumeral2. \textit{\textbf{MAMS}}

\textbf{Review}: After ordering drinks, we both decided on the Paella Vallenciana, brought out on hot plates.

\textbf{Category}: food & Neutral&The positive sentiment polarity of the review towards the aspect of food is due to the description of the Paella Vallenciana being brought out on hot plates. This suggests that the dish was served fresh and hot, enhancing the dining experience and leaving a positive impression.
\\
\hline
\romannumeral3. \textit{\textbf{COVID19}}

\textbf{Tweet}: @Timewalkproject @BBCNormanS @BBCNews A lot less palaver than being hooked up to a life support machine! 
\#WearAMask 

\textbf{Target}: Face Mask

&Favor &The tweet expresses a negative stance towards face masks, referring to them as COVID-19. The reason for this stance is not provided within the given word limit.

\\
\hline
\romannumeral4. \textit{\textbf{DEBA}}

\textbf{Sentence 1}: 
Energy mostly controls nuclear weapons and nuclear power plants, not fossil fuels. That's why they put a nuclear physicist (Muniz) in charge of it. What do people think Muniz going to do that will be harmful to the environment? Is this just an indirect anti-nuclear attack? Or just ignorance by insignificant, fringe environmental groups? 

\textbf{Sentence 2}: 
DoE also has the National Renewable Energy Lab, where a lot of the tech development used for renewable energy has happened. The DoE funded the development of fracking. There's a history of recent appointees using positions there to champion the fossil fuels industry, even if most of what the DoE does is nuclear. & Agree & Sentence 2 disagrees with sentence 1 because it suggests that governments have not used taxes for the intended purposes. This implies that they raid the education budget to raise taxes for education, contradicting the claim in sentence 1 that the purpose of a carbon tax is to.

\\
\hline
\romannumeral5. \textit{\textbf{DEBA}}

\textbf{Sentence 1}: 
I can't imagine believing Joe "Nothing Will Fundamentally Change" Biden is going to do anything except continue to serve his corporate masters.  This is...  uh, not a guy who's gonna accomplish much in office

\textbf{Sentence 2}: And that will please you, because you'd rather complain about it and have your cynicism be validated.
&Disagree & Sentence 2 agrees with Sentence 1 because it argues that the person who posts here is not trying to debate, but rather to annoy. It suggests that the solution is to simply out-annoy them and ban them.

\\
\hline 
\end{tabular}   
\vspace{-2mm}
\caption{Case study of the unreasonable explanation of our model. The model is trained on 500 labeled data instances following the AL process in Section \ref{MainResults}. }   
\vspace{-3mm}
\label{BadCase}
\end{table*}

\section{Representation Visualization}
\label{VisAPP}
\begin{figure*}[h]
    \centering
    \includegraphics[width=0.57\hsize]{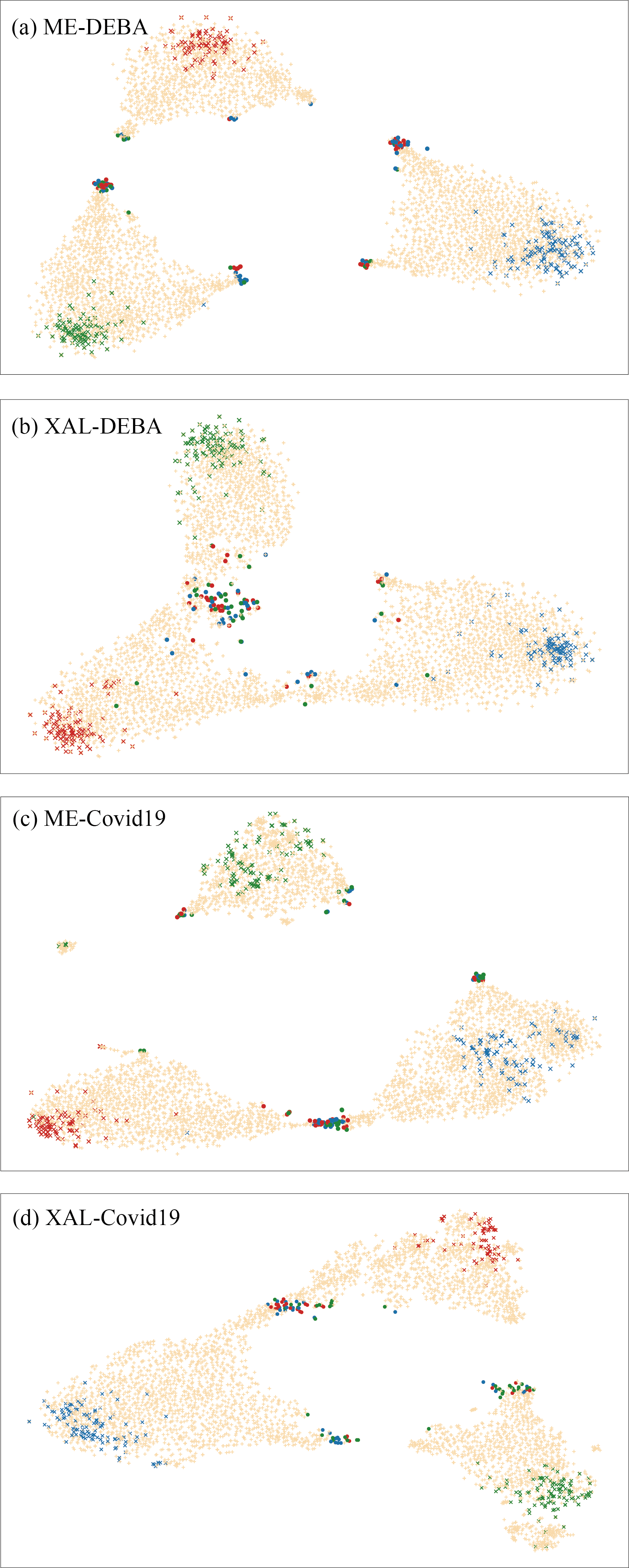}
    \caption{The t-SNE visualization of sentence representations in the data selection process.  }
    \label{visulize2}
    \vspace{-3mm}
\end{figure*}

We further demonstrate more visualizations (Figure \ref{visulize2}) in DEBA and Covid19 to show the effectivness of XAL in exploring informative data.
